\documentclass[10pt, a4paper]{article} 
\usepackage[margin=1.1in]{geometry}
\usepackage{hyperref}
\usepackage{url}
\usepackage{graphicx} 
\usepackage{subfigure} 
 \usepackage{xcolor}

\usepackage{hyperref}
\usepackage{amsthm,amsmath,amssymb,multirow,verbatim}
\usepackage{footmisc}

\usepackage{float}

\title{Semi-supervised Learning on Large Graphs:\\ is Poisson Learning a Game-Changer?}
\author{Canh Hao Nguyen}
\date{%
    Bioinformatics Center, ICR, Kyoto University\\
    Uji, Kyoto, 611-0011, Japan\\
    \today
}
\begin{document}
\maketitle

\begin{abstract}
We explain Poisson learning \cite{Calder} on graph-based semi-supervised learning to see if it could avoid the problem of global information loss problem as Laplace-based learning methods on large graphs. From our analysis, Poisson learning is simply Laplace regularization with thresholding, cannot overcome the problem.
\end{abstract}
\section{Introduction}

Given a graph $G=(V,W)$ with $V = \{x_0, \cdots x_{n-1}\}$ as a set of training data and $W$ is a $n\times n$ real nonnegative weight matrix (weights of the edges of the graphs). Supposed that we are given $m<n$ labels $y_0 \cdots y_{m-1}$ (real or binary) of $m$ nodes ($x_0\cdots x_{m-1}$) , one wishes to infer the labels of the remaining nodes ($x_{m} \cdots x_{n-1}$): $u(x_i)\ :\ V \rightarrow \mathcal{R}$ (binary class labels are usually of the form $sign(u)$). 
Graph-based models have been the main tools for semi-supervised learning.

\subsection{Laplace Learning}
Traditional Laplace learning \cite{Zhu} would use the Laplacian regularization  in this variational form:
\begin{equation}\label{LaplaceVar}
u = \arg\min u^T L u\ |\ u(x_i) = y_i, 0\leq i < m
\end{equation}
with $L = D-W$ being the graph Laplacian with $D$ being the degree diagonal matrix ($D = diag(W\times 1_n)$). The solution of (\ref{LaplaceVar})
 satisfies
 \begin{align}\label{Laplace}
\color{blue}{u(x_i)} & \color{blue}{\ = y_i \ | \ 0\leq i < m},  \\
 Lu(x_i) &= 0 \ | \  m\leq i < n.
 \end{align}
 
 \subsection{Laplace Regularization}
 
 Laplace learning can be modified to introduce a loss-based data term \cite{Nadler} as follows:
  \begin{equation}\label{relaxL}
u = \arg\min (y-u)^2 + \lambda u^TLu,
\end{equation}
with the loss $(y-u)^2$ only on labeled data points  is added to the Laplace regularization term.

\subsection{Global Information Loss Problem}
 The problem with Laplace learning, or Laplace regularization is that in large graphs, the regularization term $u^T L u$ does transfer the label from labeled nodes to far away nodes in the graphs. Concretely, the following problems are proven:
 
\begin{enumerate}
\item $u$ does not carry label information for the whole dataset, becoming a non-informative function \cite{Nadler}, unrelated to the data distribution \cite{Alaoui}. This leads to the learnt function on the graph to be "peaky" that peaks at 
the labeled points and quickly becomes close to a constant value away from the labeled point.
\item Commute time distance $ct$ does not contain graph information for fixed $x_i, x_j$ as $n \rightarrow \infty$ (for a constant $c$) \cite{Luxburg}: 
$$
ct(x_i,x_j) \rightarrow \frac{c}{d_i}+\frac{c}{d_j}.
$$
\item Laplacian kernel between data points are close to zero for fixed $x_i, x_j$ as $n \rightarrow \infty$:
$$
K(x_i,x_j) \rightarrow 0. 
$$
\end{enumerate}

There are many methods that claimed to overcome the problem without any rigorous proof. \textcolor{red}{Exceptions} are two methods with mathematically proven properties that could overcome the problems of Laplace-based learning. 
\begin{itemize}
\item $l_p$ Laplacian regularization-based distance results in a function that is most related to data distribution when $p=d+1$, with $d$ being the intrinsic dimension of the data distribution \cite{Alaoui}, overcoming the first problem. This method is computationally infeasible. 
\item $l_p$ norm-based distances computed from $L^{\dagger}$ are proven not have the second problem \cite{Nguyen}. This method is computationally feasible for medium-sized graphs.
\end{itemize}

 \subsection{Poisson Learning}
  Poisson learning \cite{Calder} is different from Laplace learning in replacing boundary condition ($u(x_i) = y_i \ | \ 0\leq i < m$) with $Lu(x_i) = c \in \mathcal{R} | \ 0\leq i < m$ . It is \emph{claimed} to avoid the peaky label function problem and let the known label propagates further compared to Laplace learning. For $\bar{y} = mean(y_1 \cdots y_m)$, the solution of Poisson learning satisfies:
  \begin{align}\label{Poisson}
 \textcolor{blue}{Lu(x_i)} &\color{blue}{\ = y_i - \bar{y} \ | \ 0\leq i < m}, \\
 Lu(x_i) &= 0 \ | \  m\leq i < n, \\
 \sum_{i= 1}^n d_iu(x_i) &= 0.
 \end{align}
 This correspond to the solution of the following variational problem: 
\begin{equation}\label{PoissonVar}
u = \arg \min \frac{1}{2}u^TL u - \sum_{i= 1}^m  (y_i-\bar{y}) u(x_i)\ |\ \sum_{i= 1}^n d_iu(x_i) = 0.
\end{equation} 
  
 Essentially, the difference between Laplace learning and Poisson learning is on the labeled nodes: the former \emph{fixes the labels} of the nodes while the latter \emph{fixes the smoothness} of labels on the nodes to some constants. Basically, in both Laplace and Poisson learning methods, the labels are propagated by minimizing $u^TL u$.
 
\section{Kernel viewpoint}
 
Let $u(X) = [v , z]$ be the function $u$  on the two parts of the data, $v = u[0\cdots m-1] \in \mathcal{R}^m$ being the labeled part of $u$ and $z = u[m\cdots n-1] \in \mathcal{R}^{n-m}$ being the unlabeled part of $u$. Let $K = L^{\dagger}$ be the  Moore-Penrose inverse of $L$, being the Laplacian kernel $K$.   In the RKHS induced by the kernel of each method, let $\phi(x_i)$ denotes the image of sample $x_i$. We show that all the above methods have decision functions of the form with a constant $c \in \mathcal{R}$, called offset, acting as classification thresholding: 
\begin{equation}\label{alldecisionfunctions}
u(x_i) = \sum_j \alpha_j K_{ij} + c = <x_i, \sum_j \alpha_j \phi{x_j}> + c. 
\end{equation}
\subsection{Laplace learning} 
 Laplace learning become: 
\begin{equation}\label{Laplace2}
u = \arg\min f(u) (= [v^T,z^T] L [v,z]).
\end{equation}
Let $L_1 = L[0\cdots m-1,0\cdots m-1]$, $L_2 = L[m\cdots n-1,m\cdots n-1]$, $L_{12 } = L[0\cdots m-1,m\cdots n-1]$ and $L_{21}= L[m\cdots n-1,0\cdots m-1]$. Then 
$$
f(u) = v^T L_1 v + 2 z^T L_{21} v + z^T L_{2}z. 
$$
Taking the derivative of $f$ on the variable part $z$ (as $v$ is the fixed part), then the first order condition becomes: 
\begin{align}
\frac{\partial f}{\partial v} = 2L_2 z + 2L_{21} v  &= 0\\
z &= -L_2^{-1} L_{21} v + c ker(L_2)
\end{align}

\textbf{Kernel representation:} Let $K_2 = L_2^{-1}$, then $K_2$ is the Laplacian kernel on the unlabeled part of the graph. 

\textbf{Weight vector $\alpha$:} $-L_{21}v$: the weight of each nodes becomes the sum of  \emph{edge weights multiplied by the labels} to the labeled nodes, i.e., only border nodes have weights. 
\begin{equation}\label{Laplaceweight}
	\alpha_i = \sum_{j=1}^m w_{ij} y_j
\end{equation}

\textbf{Offset $c$:} usually not taken into account, namely $c=0$

\subsection{Laplace regularization} 
The Laplace regularization method, sometimes called soft constraint method, has the following form:

\begin{equation}\label{relaxL}
u = \arg\min \lambda (y-u)^2 +  u^TLu.
\end{equation}
The solution is 
\begin{equation}
u = ( L + \lambda I)^{-1}y  = L^{\dagger}y +  \frac{y}{\lambda}.
\end{equation}
$\lambda$ allows for a linear interpolation between the solution of $Ky (= L^{\dagger}y)$ and $y$. With appropriate scaling of $y$ to account for class imbalance, $Ky$ can be considered as the nearest class mean classifier. 

\textbf{Kernel representation:} $K = L^{\dagger}$

\textbf{Weight vector $\alpha$:} only on labeled nodes ($j<m$), $\alpha_i = y_i$. 

\textbf{Offset $c$:} usually not taken into account, namely $c=0$

\subsection{Poisson learning}
We show the solution of the unconstrained problem satisfies the constraint.
\begin{align}
u &= \arg \min \frac{1}{2} u^TL u - \sum_{i= 1}^m  (y_i-\bar{y}) u(x_i) 
\end{align}

Taking derivative, with $t \in \mathcal{R}^n$, $t_i = y_i-\bar{y}$ for $0\leq i < m$ and $t_i = 0$ otherwise. Given that the graph is connected: 
\begin{align}\label{PoissonSol}
Lu - t &= 0 \nonumber \\
u &= L^{\dagger}t + c 1_n
\end{align}
for some $c \in \mathcal{R}$, $1_n (= ker(L))$ being the vector of all 1 in $\mathcal{R}^n$. We now prove that there exists a $c$ that satisfies the constraint.
\begin{align}
\sum_{i=1}^n d_i u(x_i) &=  <d,  L^{\dagger}t + c 1_n>\\ 
&= <d,  L^{\dagger}t> + c \sum_{i=1}^n d_i\\
\end{align}
Therefore, to have $\sum_{i=1}^n d_i u(x_i) = 0$,
\begin{align}
c = - \frac{<d,  L^{\dagger}t>}{\sum_{i=1}^n d_i}
\end{align}
This is the unique solution of Laplace learning model.

\textbf{Kernel representation:} $K = L^{\dagger}$

\textbf{Weight vector $\alpha$:} only on labeled nodes ($i < m$), $\alpha_i = y_i - \bar{y}$. 

\textbf{Offset $c$:} $c =  - \frac{<d,  L^{\dagger}t>}{\sum_{i=1}^n d_i}$.

\begin{table}
	\centering
		\begin{tabular}{|l|l|l|l|} \hline
Method & Kernel & $\alpha$ & Classification functions  \\ \hline	
 Laplace  &  $L_2^{\dagger}$ &   $-L_{21}v$ & nearest class prototype \\
 Regularization &   $L^{\dagger}$ &  $y$ & nearest class mean  \\ 
 Poisson  &  $L^{\dagger}$ &   $y-\bar{y}$ ($i<m$) & nearest class mean  \\
\hline		\end{tabular}
		\caption{Comparing decision functions of methods}\label{tab1} 
\end{table}

\subsection{Comparison} 
\begin{itemize}
\item All three methods can be interpreted as in (\ref{alldecisionfunctions}). It is different from SVMs in the sense that $\sum_{alpha_i}$ might not be equal to $0$. 
\item For Laplace learning, weight vector $\alpha$ might not sum to $0$ even with centralizing $y$ to have labels summed to $0$. This might make biased decision favoring the class with more weights to labeled nodes (such as labeled nodes of high density).
\item For Laplace regularization, weight vector $\alpha$ might not sum to $0$, but with centralizing $y$, the decision function is \emph{nearest class mean classifier}. 
\item For Poisson learning can be seen as \emph{Laplace regularization with a chosen threshold $c$}. 
\end{itemize}

\subsection{Conclusion}
\begin{enumerate}
\item What is the problem with Laplace learning? The weight vector $\alpha$ depend on edge weights adjacent to labeled nodes. Centralizing labels does not solve the problem. Solution? Centralizing $\alpha$ will make it a \emph{nearest class mean classifier}, with class-mean is a weighted sum of border nodes. 
\item Laplace regularization with label centralization becomes \emph{nearest class mean classifier}. 
\item What is Poisson learning in the RKHS? It becomes \emph{nearest class mean classifier} with an offset (or Laplace regularization with an offset). 
\item Can Poisson learning avoid the global information loss problem on large graph? The answer is $\color{red}{NO}$ due to $\L^{\dagger}$ representation. 
\item What is the advantage of Poisson learning? Offset $c$, which acts as a threshold for classification. Can it improve classification errors on Laplace regularization? Possible if the offset is meaningful. Can it improves AUC scores on Laplace regularization? No, they give the same AUC scores.
\item What is the meaning of the offset $c$? It depends on how the weights are constructed. One way to interpret is that $\sum_i sign(u_i) d_i = 0$ would be giving the two class an equal volume.  
 
\item What happened on the extremely small training sizes \cite{Calder}? Actually, it is the problem with Laplace learning, even with equal numbers of labeled points for each class,  densities (on the underlining distributions) at the sampled points may vary greatly, making $u$ unstable. Laplace regularization (with label centralization) and Poisson learning do not have this problem. More data would tend to avoid this problem as labeled data density converges to mean class density.
\end{enumerate}

\bibliographystyle{plain}
\bibliography{ssl}

\begin{thebibliography}{1}

\bibitem{Alaoui}
Ahmed~El Alaoui.
\newblock Asymptotic behavior of
  {\textbackslash}({\textbackslash}ell{\_}p{\textbackslash})-based laplacian
  regularization in semi-supervised learning.
\newblock In Vitaly Feldman, Alexander Rakhlin, and Ohad Shamir, editors, {\em
  Proceedings of the 29th Conference on Learning Theory, {COLT} 2016, New York,
  USA, June 23-26, 2016}, volume~49 of {\em {JMLR} Workshop and Conference
  Proceedings}, pages 879--906. JMLR.org, 2016.

\bibitem{Calder}
Jeff Calder, Brendan Cook, Matthew Thorpe, and Dejan Slepcev.
\newblock Poisson learning: Graph based semi-supervised learning at very low
  label rates.
\newblock In {\em Proceedings of the 37th International Conference on Machine
  Learning, {ICML} 2020, 13-18 July 2020, Virtual Event}, volume 119 of {\em
  Proceedings of Machine Learning Research}, pages 1306--1316. {PMLR}, 2020.

\bibitem{Nadler}
Boaz Nadler, Nathan Srebro, and Xueyuan Zhou.
\newblock Statistical analysis of semi-supervised learning: The limit of
  infinite unlabelled data.
\newblock In Yoshua Bengio, Dale Schuurmans, John~D. Lafferty, Christopher
  K.~I. Williams, and Aron Culotta, editors, {\em Advances in Neural
  Information Processing Systems 22: 23rd Annual Conference on Neural
  Information Processing Systems 2009. Proceedings of a meeting held 7-10
  December 2009, Vancouver, British Columbia, Canada}, pages 1330--1338. Curran
  Associates, Inc., 2009.

\bibitem{Nguyen}
Canh~Hao Nguyen and Hiroshi Mamitsuka.
\newblock New resistance distances with global information on large graphs.
\newblock In Arthur Gretton and Christian~C. Robert, editors, {\em Proceedings
  of the 19th International Conference on Artificial Intelligence and
  Statistics, {AISTATS} 2016, Cadiz, Spain, May 9-11, 2016}, volume~51 of {\em
  {JMLR} Workshop and Conference Proceedings}, pages 639--647. JMLR.org, 2016.

\bibitem{Luxburg}
Ulrike von Luxburg, Agnes Radl, and Matthias Hein.
\newblock Hitting and commute times in large random neighborhood graphs.
\newblock {\em J. Mach. Learn. Res.}, 15(1):1751--1798, 2014.

\bibitem{Zhu}
Xiaojin Zhu, Zoubin Ghahramani, and John~D. Lafferty.
\newblock Semi-supervised learning using gaussian fields and harmonic
  functions.
\newblock In Tom Fawcett and Nina Mishra, editors, {\em Machine Learning,
  Proceedings of the Twentieth International Conference {(ICML} 2003), August
  21-24, 2003, Washington, DC, {USA}}, pages 912--919. {AAAI} Press, 2003.

\end{thebibliography}

\end{document}